# Un système modulaire d'acquisition automatique de traductions à partir du Web

Stéphanie Léon

LIRMM, 161 rue Ada, 34392 Montpellier Cedex 5[1]
stephanie.leon@lirmm.fr

**Résumé** Nous présentons une méthode de Traduction Automatique d'Unités Lexicales Complexes (ULC) pour la construction de ressources bilingues français/anglais, basée sur un système modulaire qui prend en compte les propriétés linguistiques des unités sources (compositionnalité, polysémie, etc.). Notre système exploite les différentes « facettes » du Web multilingue pour valider des traductions candidates ou acquérir de nouvelles traductions. Après avoir collecté une base d'ULC en français à partir d'un corpus de pages Web, nous passons par trois phases de traduction qui s'appliquent à un cas linguistique, avec une méthode adaptée : les traductions compositionnelles non polysémiques, les traductions compositionnelles polysémiques et les traductions non compositionnelles et/ou inconnues. Notre évaluation sur un vaste échantillon d'ULC montre que l'exploitation du Web pour la traduction et la prise en compte des propriétés linguistiques au sein d'un système modulaire permet une acquisition automatique de traductions avec une excellente précision.

**Abstract** We present a method of automatic translation (French/English) of Complex Lexical Units (CLU) for aiming at extracting a bilingual lexicon. Our modular system is based on linguistic properties (compositionality, polysemy, etc.). Different aspects of the multilingual Web are used to validate candidate translations and collect new terms. We first build a French corpus of Web pages to collect CLU. Three adapted processing stages are applied for each linguistic property : compositional and non polysemous translations, compositional polysemous translations and non compositional translations. Our evaluation on a sample of CLU shows that our technique based on the Web can reach a very high precision.

**Mots-clés :** Traduction Automatique, Unités Lexicales Complexes, Désambiguïsation lexicale, World Wide Web, Terminologie

## 1 Introduction

L'ambiguïté lexicale est un problème majeur pour les systèmes de traduction automatique ou de recherche d'informations interlingue. Par exemple, la traduction anglaise du terme français

---
[1] Cet article est issu d'un travail de thèse soutenue en décembre 2008 sous la direction de Jean Véronis (LIF).

*caisse* est différente selon que l'usage concerne, entre autres, *l'INSTRUMENT DE MUSIQUE* (*drum*), la *BANQUE* (*fund*) ou la *VALISE* (*case*). L'absence de désambiguïsation lexicale pour la traduction automatique conduit à des résultats qui gênent souvent la compréhension. Ainsi, le système Systran[2] traduit l'Unité Lexicale Complexe (ULC)[3] *caisse claire* par *clear case*, ce qui est incompréhensible pour un anglophone. La co-occurrence *caisse/claire* constitue un indice désambiguïsateur très fort, qui, si elle était correctement enregistrée dans une base de données bilingue pourrait servir à générer des traductions correctes (*caisse claire > snare drum*). La polysémie est rendue faible dès que l'on envisage les unités lexicales selon leur co-occurrent (Yarowsky, 1993), (Shütze, 1998) et (Véronis, 2003). Les travaux existants ont proposé des méthodes d'acquisition de terminologie bilingue à partir de corpus parallèles (voir (Véronis, 2000)) ou comparables (citons, entre autres, (Rapp, 1999), (Fung, McKeown, 1997) (Fung, Yee, 1998) et (Morin *et al.*, 2004)). Les corpus parallèles constituent des ressources rares tandis que les corpus comparables se limitent à un domaine de spécialité. Le Web, qui génère des besoins considérables en traduction, offre en même temps un réservoir gigantesque de données multilingues qui pourraient être exploitées afin d'acquérir les traductions correctes.

Les contraintes de Traduction Automatique d'ULC varient en fonction des caractéristiques linguistiques des unités sources. Si les constituants de l'ULC sont polysémiques, la tâche consiste à sélectionner la bonne traduction de chaque constituant parmi des traductions candidates. Si les constituants ne sont pas polysémiques, la tâche consiste à valider ou pas la combinaison de leurs traductions. Les traductions d'ULC peuvent être compositionnelles, c'est-à-dire basées sur une combinaison des traductions des constituants, ou non compositionnelles (traduction non littérale). Les travaux d'acquisition de traductions n'ont pas exploité ces indices linguistiques et ont proposé des méthodes globales. Nous appliquons un traitement automatique adapté à une typologie linguistique, pour la construction d'un système modulaire d'acquisition automatique de traductions d'ULC du français vers l'anglais, à partir du Web. Ce système s'appuie sur la détection automatique des propriétés linguistiques de traduction des ULC (compositionnalité, polysémie des constituants) et chaque module est consacré à un type de traduction. Les propriétés multilingues du Web sont exploitées afin de collecter et de filtrer automatiquement des traductions candidates (pages parallèles ou « partiellement » parallèles, fréquences, comparaison de mondes lexicaux). Notre système permet d'acquérir des ULC sources en français, à partir d'un corpus de pages Web et de proposer un système modulaire de traductions de ces ULC. La section 2 présente notre méthodologie, l'acquisition automatique des données et la méthode de traduction sur un échantillon aléatoire. La section 3 présente l'évaluation, les résultats et les perspectives.

## 2  Méthodologie et traitement des données

Nous procédons à une extraction d'ULC en français à partir d'un corpus de pages Web étiqueté morpho-syntaxiquement. Nous analysons automatiquement leurs propriétés linguistiques via un dictionnaire électronique bilingue. Le module de traduction est sélectionné en fonction de ces caractéristiques. Après avoir décrit la méthode d'acquisition d'ULC sources (2.1), nous décrivons l'architecture du système et détaillons chaque module (2.2).

---

[2]  http://www.systransoft.com/

[3]  Nous parlons d'ULC afin de désigner une cooccurrence lexicale entre deux lexèmes liés syntaxiquement.

*Un système modulaire d'acquisition automatique de traductions à partir du Web*

## 2.1 Acquisition d'ULC en langue source

Nous collectons des pages Web associées à une liste aléatoire de noms simples extraits du dictionnaire électronique bilingue *Collins Pocket* (français-anglais)[4]. Ces pages sont nettoyées automatiquement et étiquetées via le logiciel *Treetagger*[5]. Nous appliquons ensuite des filtres linguistiques et des filtres de fréquence. D'un point de vue linguistique, nous définissons les patrons morpho-syntaxiques répondant aux relations de dépendances syntaxiques recherchées (NOM-ADJECTIF ET NOM-de(d')-NOM) en ne prenant en compte que les séquences contigües. En ce qui concerne la fréquence, nous posons des seuils fixes. La fréquence de l'ULC au sein de notre corpus doit être supérieure ou égale à 10. Ensuite, les ULC sont testées en tant que requêtes sur le moteur de recherche *Yahoo* et nous ne conservons que celles dont les nombres d'occurrences estimés par le moteur de recherche sont les plus élevés [6]. Ce filtre peut provoquer du silence, mais nous posons volontairement un filtre élevé afin d'obtenir des ressources de très bonne qualité, sans aucune intervention humaine. Cette étape pourra être améliorée par la suite par des méthodes statistiques plus poussées, mais elle constitue en l'état un banc de test satisfaisant afin d'évaluer notre système de traduction. Nous obtenons 9664 ULC en français, associées à 1664 têtes sémantiques[7] (noms simples dont vont dépendre sémantiquement les autres actants au sein d'une ULC). Le nombre moyen d'ULC par tête sémantique est de 6. Notre corpus de pages Web est constitué d'environ 4 160 000 pages Web et continue de grossir au quotidien. Le schéma suivant présente le nombre d'ULC sources obtenus par patron morpho-syntaxique :

| NOM-ADJECTIF | 5 166 |
|---|---|
| NOM-DE-NOM | 2934 |
| NOM-D'-NOM | 1564 |
| TOTAL | 9664 |

Tableau 1 : Proportion d'ULC par patron morpho-syntaxique

Voici un extrait d'ULC associées à la tête sémantique *appareil* :

*NOM-D'-NOM : Appareil d'état, appareil d'imagerie, catégorie d'appareil*
*NOM-DE-NOM : Appareil de chauffage, appareil de contrôle, appareil de cuisson*
*NOM-ADJECTIF : Appareil administratif, appareil argentique, appareil auditif*

Afin de tester notre méthodologie de traduction, nous réalisons un échantillon aléatoire de 1075 ULC. Nous avons évalué la qualité des ULC de façon manuelle et aucune n'a été éliminée.

---

[4] Nous aurions pu partir d'une liste monolingue d'ULC déjà existante mais il n'existe pas de ressource à très vaste échelle.

[5] http://www.ims.uni-stuttgart.de/projekte/corplex/TreeTagger/

[6] La fréquence de l'ULC retournée par le moteur de recherche en tant qu'expression littérale doit être supérieure ou égale à 10000 et la fréquence de l'ULC, précédée par un article (défini ou indéfini) doit être supérieure ou égale à 1000.

[7] Nous parlons de *tête sémantique* afin de désigner l'élément dont dépendent sémantiquement les autres actants au sein d'une ULC.

## 2.2   Architecture et spécification du système d'acquisition des traductions

Notre système est modulaire (chaque phase traite d'ULC ayant des caractéristiques spécifiques de traduction) et linéaire (les traductions non obtenues au sein d'un module sont renvoyées au module suivant). Chaque sous-section présente un module : les traductions compositionnelles non polysémiques (2.2.1), les traductions compositionnelles polysémiques (2.2.2) et les traductions non compositionnelles et/ou inconnues (2.2.3).

### *2.2.1   Traductions compositionnelles non polysémiques*

Le premier module traite des traductions compositionnelles non polysémiques : le sens est transparent (chaque constituant peut être traduit de façon littérale) et le sens de chaque constituant n'est pas ambigu (non polysémique)[8]. Cette tâche peut consister à déterminer si la combinaison de chaque traduction candidate des constituants de l'ULC, générée par une ressource dictionnairique est valide ou pas. (Léon et Millon, 2005) montrent que la fréquence sur le Web de traductions candidates peut être exploitée pour la validation si les unités sources ne sont pas polysémiques. Nous nous appuyons sur le nombre de traductions candidates de chaque constituant dans notre dictionnaire bilingue (Dagan *et al.*, 1991) et ne conservons que celles dont les constituants ne comptent qu'une traduction. Par exemple, l'ULC *ambiance musicale* est conservée puisque *ambiance* et *musicale* ne comptent respectivement qu'une seule traduction (*atmosphere*/*musical*). Nous générons automatiquement toutes les traductions candidates via le *Collins Pocket* selon la méthode de (Léon, Millon, 2005 ; Léon, 2006), qui consiste à générer toutes les combinaisons possibles des traductions des éléments simples et à effectuer des règles de transformation morpho-syntaxique, comme *institut de psychologie* (*institut / institute*, *psychologie / psychology*), dont les traductions candidates sont *institute of psychology* et *psychology institute*. Les règles de transformation morpho-syntaxiques pour chaque patron sont les suivantes :

*NOM1 de NOM 2 > NOM1 of NOM2*
*NOM1 de NOM 2 > NOM2 NOM1*
*NOM ADJECTIF > ADJECTIF NOM*

Le moteur de recherche Yahoo est interrogé automatiquement afin de récupérer le nombre d'occurrences de chaque traduction candidate. Pour chaque traduction candidate, nous générons un ensemble de requêtes selon le modèle suivant :

| **NOM-ADJECTIF** | « the ADJ NOM » OR « a ADJ NOM » |
|---|---|
| **NOM1-DE(D')-NOM2** | « the NOM1 of NOM2 » OR « a NOM1 of NOM2 » |
| | « the NOM2 NOM1 » OR « a NOM2 NOM1 » |

Tableau 2 : Patrons des requêtes des traductions candidates

Un filtre simple est appliqué aux traductions candidates. Nous ne conservons que celles dont la fréquence sur le Web est au moins égale à un dix-millième des occurrences du mot cible.

---

[8] Nous considérons qu'une unité est non polysémique si elle a strictement une unique traduction dans notre dictionnaire bilingue. Le terme de « polysémie » est donc ici employé de façon pratique, dans le contexte de la traduction et à partir d'une ressource donnée.

Prenons pour exemple[9] « *messe de minuit* » et deux de ses traductions candidates « *midnight mass* » et « *mass of midnight* » : Seuil_mass : 764 000 000 / 10 000 = 30400. La traduction « *midnight mass* » (avec une fréquence de 336 000, donc supérieure au seuil limite pour le nom cible *mass*) est retenue, tandis que « *mass of midnight* » (avec une fréquence de 65, donc inférieure au seuil limite pour le nom cible *mass*) est rejetée. Après le filtre automatique sur les fréquences, 39.24% des traductions candidates sont conservées. Les traductions obtenues comptent pour 9.11% de nos unités lexicales à traduire[10]. Parmi les unités lexicales non polysémiques, 57.40% d'entre elles obtiennent une traduction à cette étape. Le tableau suivant présente la proportion de traductions conservées après le filtre automatique.

|  | Traductions générées | *Filtre automatique* Filtre seuil fréquence | |
|---|---|---|---|
| **NOM ADJ** | 29 | 20 | 68,97% |
| **NOM DE NOM** | 40 | 8 | 20,00% |
| **NOM D' NOM** | 10 | 3 | 30,00% |
| **TOTAL** | 79 | 31 | 39,24% |

Tableau 3 : Résultats du filtre des traductions après la phase 1

### *2.2.2 Traductions compositionnelles polysémiques*

Lorsque les constituants à traduire sont polysémiques, la tâche de traduction ne consiste pas en une simple validation mais en une sélection parmi de nombreuses traductions candidates. Il s'agit de désambiguïser le sens de chaque constituant. (Léon, 2006) montre qu'une comparaison des mondes lexicaux sur le Web entre l'unité source et chaque traduction candidate permet de lever un grand nombre d'ambiguïtés lexicales. Selon nous, un « monde lexical » désigne les co-occurrences fréquentes (plus larges que le co-occurrent immédiat) d'une unité lexicale au sein d'une collection de textes (Véronis, 2003). Ce deuxième module est une version améliorée de la méthode de (Léon, 2006) qui procède à une désambigüisation lexicale pour la traduction via les contextes lexicaux sources et cibles sur le Web. Les traductions candidates sont générées avec la même méthode que celle décrite en 2.2.1. Comme celles-ci sont en grande quantité, il serait coûteux de générer les mondes lexicaux de chacune d'elle. Nous utilisons deux filtres préalables qui permettent d'éliminer les traductions les plus erronées. Nous testons les fréquences des couples de traduction sur le Web. Notre hypothèse est que si la traduction est correcte, le couple doit apparaître au moins une fois dans un même document : de nombreux documents du Web sont des textes linguistiquement mixtes (traductions de manuels, de catalogues, glossaires, etc.). Nous engendrons automatiquement une requête pour chaque couple restant, du type de « *caisse centrale* » «*central fund* » et ne conservons que les couples dont la fréquence est supérieure ou égale à 1. Notre second filtre détermine le rapport entre la fréquence sur le Web de l'ULC source et celui des traductions candidates. Par exemple, « *caisse de retraite* » apparaît 157000 fois. « *retirement cas* » apparaît 2850 fois, tandis que « *retirement fund* » apparaît 1240000 fois. On exclut les traductions ayant une fréquence inférieure au terme français (étant donné le rapport entre le nombre de pages indexées en anglais et en français, une traduction correcte en anglais doit avoir une fréquence supérieure à son ULC source en français).

---

[9] Juillet 2008.

[10] Viennent s'ajouter les ULC déjà traduites au sein de notre dictionnaire que nous conservons d'emblée, à savoir 98 traductions à l'issue de la phase 1 sur les 1075 ULC de départ.

Nous construisons enfin les mondes lexicaux de chaque ULC source et des traductions candidates restantes. Nous collectons les 1000 premiers résumés retournés par le moteur de recherche *Yahoo* pour chaque requête correspondant à une ULC source et à chacune de ses traductions candidates. Nous collectons ensuite les cinquante noms et les cinquante adjectifs les plus fréquents parmi les résumés étiquetés de façon morpho-syntaxique. Ces mondes lexicaux français et anglais pour les noms et pour les adjectifs sont respectivement comparés via le dictionnaire bilingue. Le nombre de mots communs entre les mondes lexicaux français et anglais est comptabilisé. Le coefficient de Jaquard est utilisé afin de mesurer le degré de similitude entre les deux ensembles : | *inter(X,Y)* | / | *union (X,Y)* |. Seules les traductions candidates dont l'indice de Jaquard est supérieur à un seuil fixé (pour la catégorie des noms et pour la catégorie des adjectifs) sont conservées et nous sélectionnons celle qui a le plus haut score pour chaque ULC. A l'issue de cette étape, nous obtenons pour 68.98% des ULC restantes traduire.

| Unités lexicales restantes après la phase 1 | Traductions candidates générées | *Filtre automatique* | | |
|---|---|---|---|---|
| | | Filtre Web parallèle, « top 3 » | Filtre rapport français/ anglais | Filtre indice de Jaquard |
| 977 | 18 844 | 1919 | 1239 | 674 |

Tableau 4 : Résultats des filtres après la phase 2

### 2.2.3  Traductions non compositionnelles et/ou inconnues

A ce stade, plusieurs difficultés expliquent l'absence de traduction des ULC restantes. Il arrive que la somme des traductions de chaque élément de l'ULC ne permette pas d'obtenir la traduction adéquate, comme dans *caisse claire > snare drum* (*tambour/ piège*). Il peut aussi s'agir de cas où la base et/ou le co-occurrent est recensé dans notre dictionnaire, mais l'emploi pertinent n'est pas répertorié, comme pour *caisse d'épargne > savings bank*. Ici, l'emploi de *caisse* (BANQUE) n'est pas répertorié dans notre dictionnaire. Enfin, il peut s'agir de cas où la base ou le co-occurrent est un terme technique non recensé dans nos ressources dictionnairiques comme dans *appareil **circulatoire**.* Dans cet exemple, la traduction de *circulatoire* est inconnue de nos ressources et notre objectif est également de découvrir de nouvelles traductions. Notre dernière phase vise à pallier deux difficultés, le problème de la non-compositionnalité et celui de co-occurrents inconnus de nos ressources dictionnairiques, parce qu'ils sont trop techniques ou récents. Le point commun est qu'une utilisation de ressources existantes n'est pas satisfaisante. Le principe de ce module est de collecter directement les traductions à partir de résumés « mixtes » sur le Web, à l'instar de Nagata (2001) pour le japonais et l'anglais. Notre méthode d'acquisition de résumés mixtes consiste à générer des requêtes en français (langue source), recherchées dans des pages en anglais (langue cible), ce qui ramène majoritairement des pages linguistiquement « mixtes ». A partir de ces résumés, nous appliquons deux stratégies de repérage des traductions candidates : dans un premier temps, nous identifions des cognats *(« occurrences qui sont identiques ou se ressemblent graphiquement »*, Véronis, 2000), et dans un second temps, nous repérons les couples les plus fréquents. Ces deux étapes se présentent de la même façon que les étapes précédentes, c'est-à-dire qu'elles sont successives : nous recherchons d'abord les cognats. Il peut s'agir, par exemple de mots graphiquement apparentés tels que *langue* et *language* (*ibid.*)

ou alors de formes communes. Nous comparons les quatre premières lettres du co-occurrent anglais (premier constituant du couple) avec celui du co-occurrent français (deuxième constituant). Nous aurions pu appliquer une méthode de comparaison plus classique telle qu'une distance de Levenshtein mais nous ne nous intéressons qu'aux chaines communes qui se trouvent dans la même position au sein des mots source et cible. Nous passons ensuite par plusieurs filtres de validation, selon le même modèle que dans la phase précédente (requêtes en couple sur le Web et filtre du rapport entre les fréquences françaises et anglaises). Il reste, après tous les filtres, 292 traductions candidates. Les traductions candidates restantes sont ensuite filtrées par l'indice de Jaquard. A l'issue de cette sous-étape, 89 traductions sont validées, soit 29.37% des unités lexicales de départ pour cette phase, et 8.27% de la totalité de nos données de départ. Les traductions non obtenues passent ensuite par le module des couples fréquents. Etant donné que le texte contient plusieurs langues, il est délicat de procéder à un étiquetage morpho-syntaxique. Nous conservons les résumés bruts retournés par le moteur de recherche et récoltons les couples de mots les plus fréquents au sein de ces résumés. Nous nous centrons volontairement sur les couples candidats, et ne prenons pas en compte les chaines plus longues, ce qui peut provoquer des cas de silence pour le repérage du patron *NOM-of-NOM* ou d'une unité simple. Toutefois, l'analyse de textes non étiquetés est une tâche délicate et nous nous centrons sur le patron morpho-syntaxique candidat le plus fréquent. A partir des 303 unités lexicales sources restantes à traduire, 327 815 couples différents sont générés. A l'issue des mêmes filtres que précédemment, 26 traductions sont validées (soit 115 en tout pour la phase complète). Le tableau 5 montre un exemple de traductions obtenues à chaque étape (la totalité des traductions est de 887) et le tableau 6 présente la quantité de traductions obtenues par phase.

| PHASE 1 (Méthode des fréquences) | psychologie sociale/social psychology<br>drame musical/musical drama |
|---|---|
| PHASE 2 (Mondes lexicaux) | accident grave/serious accident<br>éclat naturel/natural shine |
| PHASE 3 (Cognats et couples fréquents) | acide nucléique/nucleic acid<br>souris d'agneau/lamb shank |

Tableau 5 : Exemples de traductions

| PHASE 1 (Méthode des fréquences) | 98 | 11,05% |
|---|---|---|
| PHASE 2 (Mondes lexicaux) | 674 | 75,99% |
| PHASE 3 (Cognats et couples fréquents) | 115 | 12,97% |

Tableau 6 : Traductions obtenues pour chaque phase[11]

# 3   Evaluation et analyse des résultats

Au sein de notre échantillon aléatoire, nous évaluons les 887 traductions obtenues. Nous avons opté pour une évaluation manuelle, effectuée par un locuteur bilingue[12], avec trois appréciations possibles : A (Bonne traduction), B (Traduction acceptable), ou C (Mauvaise traduction). Le locuteur a eu accès aux contextes des traductions, en observant les contextes retournés par le moteur de recherche, pour chaque requête d'ULC source et chaque traduction obtenue. Les résultats obtenus montrent que 89,29% des traductions ont été considérées comme correctes par l'évaluatrice (catégorie A) et 5,07% ont été considérées comme acceptables (B). Seulement 5,64% de traductions ont été jugées erronées (C). Ces résultats sont particulièrement satisfaisants puisqu'ils montrent que 94,36% des résultats sont directement exploitables, sans aucune intervention humaine, soit une précision très satisfaisante. Le taux de rappel, lui, est de 77,86%. Nous analysons les erreurs de traduction (3.1), puis nous présentons nos perspectives d'évolution (3.2).

---

[11]   La phase 1 combine les traductions obtenues à cette étape avec les ULC déjà traduites dans notre dictionnaire que nous conservons telles quelles.

[12]   L'évaluatrice est Amanda Grey, traductrice professionnelle (http://www.amandagrey.com/).

## 3.1 Analyse des erreurs

Les erreurs que nous analysons ici concernent exclusivement celles qui ont été relevées via la non acceptation de l'évaluation de l'expert, à savoir le bruit. Nous recensons trois grandes catégories d'erreurs, les erreurs lexicales (3.1.1), les erreurs morpho-syntaxiques (3.1.2) et les erreurs « idiomatiques » (3.1.3).

### *3.1.1 Erreurs lexicales*

Les erreurs lexicales désignent un mauvais choix lexical d'au moins un des constituants de l'ULC. Le choix lexical peut être proche d'un point de vue thématique mais non équivalent (il s'agit de la tête sémantique, du co-occurrent ou de la totalité des éléments) comme dans l'exemple de *Parc nucléaire > nuclear energy*. Le choix lexical peut être totalement erroné, c'est-à-dire que la désambiguïsation lexicale n'a pas été correctement effectuée (il s'agit systématiquement d'un mauvais choix de la tête sémantique) comme dans **Fond d'aide > help back**.

### *3.1.2 Erreurs morpho-syntaxiques*

Un autre type d'erreur concerne les cas de traductions ayant une structure morpho-syntaxique erronée. Les erreurs morpho-syntaxiques altèrent moins la compréhension globale que les erreurs lexicales, comme dans l'exemple de *Analyse de marché > analysis of market* (au lieu de *market analysis*). Parmi les erreurs morpho-syntaxiques, nous distinguons deux cas pour la structure syntaxique source *NOM-DE-NOM* (du type de *analyse de marché*). Certaines erreurs consistent en un mauvais choix entre les structures de type « germanique » (*NOM-NOM*) et de type « roman » (*NOM OF NOM*) (Chuquet et Paillard, 1987). D'autres erreurs consistent en une non prise en compte de la structure de type « possessif », faisant intervenir le génitif.

### *3.1.3 Erreurs idiomatiques*

Un dernier type d'erreur concerne un choix lexical sémantiquement pertinent, mais dont le caractère idiomatique n'est pas pleinement satisfaisant, comme dans l'exemple *de Fête d'anniversaire > anniversary party*. Bien que cette traduction soit considérée comme acceptable et reste compréhensible, le choix lexical de *anniversary* (au lieu de *birthday*) ne correspond pas au choix le plus pertinent d'un point de vue idiomatique.

### *3.1.4 Perspectives*

#### *3.1.4.1 Thématiques de recherche*

Nous avons fait le choix de n'extraire qu'une seule traduction par unité lexicale source. Nous pourrions nous intéresser à un recensement de toutes les traductions possibles pour une ULC. Nous pourrions nous intéresser à un ensemble de domaines ou de thématiques sur le Web. Par exemple, une traduction satisfaisante en langue générale (l'usage le plus courant) peut être inadéquat dans un domaine de spécialité. Considérons la traduction *appareil numérique > digital camera*. Bien que l'usage le plus courant soit l'usage *PHOTOGRAPHIE*, la traduction *digital camera* est inappropriée dans certains domaines. Dans le domaine médical, la traduction est *digital device*. Une évolution sera de nous intéresser aux domaines de spécialité sur le Web, afin de pallier les limites liées à l'ambigüité lexicale.

#### *3.1.4.2 Amélioration de la conparaison des mondes lexicaux*

Une limite de la mesure de Jaquard (comparaison des mondes lexicaux) est de ne pas prendre en compte les différences de fréquences. Si le partage des fréquences est inégal, la comparaison est moins efficace que si la répartition des fréquences était équilibrée. Nous pourrions prendre en compte le poids (fréquence) des unités lexicales au sein de chaque monde lexical et obtenir une comparaison pondérée. De nombreuses méthodes peuvent être envisagées afin de tenir compte de la fréquence des unités au sein des textes à comparer. Dans une discussion sur le choix d'une méthode de mesure de comparaison entre deux textes, Brunet (2003) montre que quelle que soit la méthode, les différences sont peu sensibles.

#### *3.1.4.3 Ajout de ressources externes*

Prince et Chauché (2008) présentent une méthode de traduction basée sur l'exploitation de ressources de type ontologique. Chaque thesaurus en anglais (*English Roget Thesaurus*) et en français (*Thesaurus Larousse*) est exploité tel un espace vectoriel, dans lequel les entrées monolingues forment un vecteur de concepts associés. Les entrées françaises sont représentées sous la forme de leur équivalence dans l'espace anglais. La tâche de désambigüisation lexicale consiste à sélectionner le vecteur approprié au sein des vecteurs bilingues par comparaison avec un vecteur contextuel de la phrase source. L'exploitation de ressources externes pour la désambigüisation lexicale pourrait être combinée à notre méthode, afin de mêler des connaissances encyclopédiques (thésaurus) à des connaissances textuelles (mondes lexicaux).

## 4   Conclusion

Notre travail a porté sur l'analyse de l'aspect interlingue des ULC pour l'acquisition automatique de traductions à partir du Web. La prise en compte des caractéristiques linguistiques de traduction permet d'apporter un traitement adapté à chaque cas. Nous avons proposé une application de l'utilisation du Web dans le cadre d'applications linguistiques, en proposant une méthode « mixte » de stratégies. Les ressources que nous avons collectées jusqu'à présent sont de bonne qualité, avec une précision de traduction très satisfaisante, à savoir environ 94% de traductions acceptables. Le rappel est également honorable, avec un taux d'environ 77%. Nous nous sommes également centrée sur l'étude du contexte des ULC en testant ce phénomène à vaste échelle, en collectant les mondes lexicaux directement à partir du Web. Ces mondes lexicaux, en français et en anglais, ont été exploités pour la désambigüisation lexicale pour la traduction mais pourront être exploitées pour la construction de ressources de type ontologiques.

# Références